%% file: main_paper.tex
\documentclass[twoside]{article}

\usepackage[round]{natbib}

\usepackage[accepted]{styles/aistats2024}
\usepackage[utf8]{inputenc} 
\usepackage[T1]{fontenc}    
\usepackage{hyperref}       
\usepackage{url}            
\usepackage{booktabs}       
\usepackage{amsfonts}       
\usepackage{nicefrac}       
\usepackage{microtype}      

\usepackage[english]{babel}
\usepackage[utf8]{inputenc}
\usepackage{fancyhdr}
\usepackage{afterpage}
\usepackage{algorithm}
\usepackage{amsfonts}
\usepackage{booktabs}

\usepackage{siunitx}
\usepackage{algpseudocode}
\usepackage{booktabs}
\usepackage{hyperref}
\usepackage{tabularx}
\usepackage{tabulary}
\usepackage{arydshln}
\usepackage{enumitem}
\usepackage{amssymb}
\usepackage{tikz}
\usepackage{amsmath,multirow}
\usetikzlibrary{matrix}
\usepackage[capitalize]{cleveref}
\usepackage{bbm}
\usepackage{xspace}

\usepackage[framemethod=tikz]{mdframed}
\mdfdefinestyle{proofstyle}{%
linecolor=black,middlelinewidth=2pt,%
frametitlerule=true,%
apptotikzsetting={\tikzset{mdfframetitlebackground/.append style={%
shade,left color=white, right color=blue!20}}},
frametitlerulecolor=black!60,
frametitlerulewidth=1pt,
innertopmargin=\topskip,
}
\mdtheorem[style=proofstyle]{proof}{Proof}
\mdtheorem[style=proofstyle]{definition}{Definition}
\mdtheorem[style=proofstyle]{condition}{Condition}
\mdtheorem[style=proofstyle]{goal}{Goal}

\DeclareMathOperator*{\argmax}{arg\,max}
\DeclareMathOperator*{\argmin}{arg\,min}

\newcommand{\sep}{\!\;|\;\!}

\newcommand{\E}{\mathbb{E}}

\newcommand{\ind}{\mathbbm{1}}

\newcommand{\xfs}{{\scshape Fix-FinExp}\xspace}
\newcommand{\xis}{{\scshape Fix-InfExp}\xspace}
\newcommand{\fs}{{\scshape FinExp}\xspace}
\newcommand{\is}{{\scshape InfExp}\xspace}
\newcommand{\fx}{{\scshape Fixed}\xspace}

\newcommand{\acc}{\text{\scshape Acc}\xspace}

\begin{document}
\twocolumn[

\aistatstitle{Bayesian Online Learning for Consensus Prediction}

\aistatsauthor{ Sam Showalter$^{* \ 1}$ \And Alex Boyd$^{* \ 2}$ \And  Padhraic Smyth$^{1,2}$ \And Mark Steyvers$^{1,3}$ }

\aistatsaddress{ $^{1}$Department of Computer Science \And  $^{2}$Department of Statistics \\ University of California, Irvine \And $^{3}$Department of Cognitive Science  } ]

\input{sections/abstract}

\input{sections/introduction}

\input{sections/related_work}

\input{sections/methodology}

\input{sections/experiments}

\input{sections/discussion}

\input{sections/acknowledgements}

\bibliography{citations/citations,citations/gavin_neurips,citations/notes_citations,citations/human_committee}

\newpage

\appendix
\onecolumn
\aistatstitle{Appendix: Bayesian Online Learning for Consensus Prediction}
\input{appendix/limiting_case}

\input{appendix/experimental_setup}

\input{appendix/additional_experiments}

\end{document}

%% file: sections/abstract.tex

\begin{abstract}
Given a pre-trained classifier and multiple human experts, we investigate the task of online classification where model predictions are provided for free but querying humans incurs a cost. In this practical but under-explored setting, oracle ground truth is not available. Instead, the prediction target is defined as the consensus vote of all experts. Given that querying full consensus can be costly, we propose a general framework for online Bayesian consensus estimation, leveraging properties of the multivariate hypergeometric distribution. Based on this framework, we propose a family of methods that dynamically estimate expert consensus from partial feedback by producing a posterior over expert and model beliefs. Analyzing this posterior induces an interpretable trade-off between querying cost and classification performance. We demonstrate the efficacy of our framework against a variety of baselines on CIFAR-10H and ImageNet-16H, two large-scale crowdsourced datasets. 
\end{abstract}

%% file: sections/introduction.tex

\section{INTRODUCTION}

As machine learning classifiers have become increasingly  part of society over the past decade, there is a growing interest in integrating the decisions of humans and machines, particularly for high-stakes applications such as medicine and autonomous vehicles. For example,  instead of using a fully automated (or manual) system to classify  medical images, a preferred approach may be to leverage the strengths of both models and human experts, rather than solely relying on one or the other. A number of different problems have been investigated in this context, including frameworks for learning when to defer to a model \citep{madras2017predict,mozannar2020consistent}, learning when to switch between human and AI agents or to delegate tasks \citep{lubars2019ask,meresht2022learning,pinski2023ai}, and learning how to combine predictions from both humans and models \citep{steyvers2022bayesian}.

Distinct from prior work, a key aspect of our scenario of interest is that the target variable $y$ is defined as the consensus (plurality) vote of $N$ human experts,  rather than assuming that $y$ is available via %
an oracle (e.g. a single, infallible expert or a direct measurement of $y$).  We use pre-trained model predictions, along with a subset of expert votes, to predict expert consensus as if all experts had been queried. Like many active and online learning settings, our goal is to maximize our classification accuracy (of consensus) given an annotation budget.

\footnotetext[0\def\thefootnote{}]{* Authors contributed equally; correspondence to \texttt{showalte@uci.edu}} 

In particular, assume we are required to make online class label predictions for examples $x_t$ (e.g., medical or astronomical images)  over time $t = 1, 2, \ldots, T$, where the examples $x_t$ and labels $y_t \in \{1,\ldots,K\}$ are being drawn IID from some unknown distribution $\mathbb{P}(x,y)$. To make a prediction for each $x_t$,  we have access to $K$ class probabilities produced by a pre-trained model $f$ at no cost. In addition, we have access to label prediction ``votes'' from $N$ human experts at some cost per expert; we can query no experts, one expert, two experts, and etc., up to $N$ total experts. We assume that the accuracy of the pre-trained model and the identity of individual experts are unknown. Instead, we utilize model predictions and partially observed collections of expert votes. 
The key problem of interest is how to optimize predictive performance while trading-off the model's predictions (at zero cost) with the human expert votes (at non-zero cost) on a per-sample basis.  For example,  if $N$ is odd we identify consensus by querying until one class possesses $(N+1)/2$ expert votes.
However, in practice we may be able to query fewer experts and still accurately estimate consensus, depending on the quality of the model's predictions.

This type of problem is relatively common in practice, where the best we can do to determine ground truth is rely on the consensus of human experts, e.g., in citizen science \citep{wright2017transient,beck2018integrating} or in medicine \citep{bien2018deep,rajpurkar2020chexaid,stutz2023evaluating}. For example, a hospital may have access to a pre-trained model $f$ as well as its own set of expert radiologists. The hospital views the consensus of the expert radiologists as the ideal predictor, but would like to have an algorithmic workflow where the expert radiologists are consulted only to the extent that they are needed, depending on model predictions.

We adopt a Bayesian approach to this problem, using a generative model for  model predictions and expert votes. Our Bayesian framework allows us to update our beliefs sequentially as examples $x_t$, model predictions $f_t$, and expert votes are observed. We show how the model's predictions and observed expert votes can be used as a prior to drive beliefs about unobserved expert votes, as well as how we can learn the parameters of the prior in an online manner. This framework allows us to account for uncertainty in (i) deciding whether or not to query experts given an example $x_t$, as well as in (ii) making final classification predictions for $y_t$.

In summary, our primary contributions are as follows:
\begin{itemize}[itemsep=0pt, leftmargin=12pt, topsep=0pt]
    \item We develop a general Bayesian framework for online learning of human expert consensus, based on a multivariate hypergeometric likelihood model.
    \item We identify a computationally simple limiting case of the multivariate hypergeometric approach for when the number of experts becomes infinitely large.
    \item We propose a well-justified heuristic tied to the model's predictive beliefs on expert consensus to decide when to query and when to predict.
    \item We systematically evaluate our methods on two large, human-annotated datasets and demonstrate that our proposed approach is significantly more efficient than competing baselines for this problem.
\end{itemize}
Code implementations of methods and baselines, and experiment scripts, will be  made publicly available.

%% file: sections/related_work.tex

\section{RELATED WORK} 
Our work is broadly related to concepts in crowd-sourcing, human-AI collaboration, and active online learning and model selection. As we discuss below, what distinguishes us from  prior work is our focus on (i) predicting the consensus of a set of experts (rather than an oracle ground truth), and (ii) the performance-cost trade-offs that result from being able to query a variable number of experts per example conditioned on model predictions.

The general idea of using multiple human annotators to assign class labels, rather than a single annotator or oracle, has a long history in machine learning in the context of crowd-sourcing and citizen science \citep{beck2018integrating,sheng2019machine,uma2021learning,sayin2021review}. Work on analyzing human consensus goes back at least to Cohen's proposal of the Kappa agreement statistic \citep{cohen1960coefficient}, evolving in modern times to analysis of human labeling in applications  such as natural language processing \citep{plank-2022-problem}, %
and medical diagnosis \citep{stutz2023evaluating}. Techniques such as active online learning  have also been explored in this context, where the focus is on learning a model $f$ with minimal human supervision. In this general setting, the most relevant body of prior work involves the use of model predictions to assist the human labeling process  \citep{wright2019help}. A commonly used technique in this context is the use of confusion matrices to infer patterns of labeling errors for individual humans and machines \citep{branson2017lean,van2018lean,tanno2019learning,kerrigan2021combining}. However, 
inherent in this work is the assumption that annotators are noisy estimators of a separate oracle ground truth $y$. By contrast, we specifically assume the consensus of human experts solely defines the label $y$ that we wish to predict.  

\paragraph{Human-AI Collaboration} A closely-related and recent body of work in machine learning develops algorithmic approaches for various workflows in human-AI collaboration (e.g., see \cite{jarrett2022online}), such as algorithms that allow models to ``learn to defer'' to human experts \citep{madras2017predict,mozannar2020consistent,verma2022calibrated}. Of particular relevance is work that focuses on optimally combining model and expert predictions, in the situation where both are assumed to be available (e.g., \cite{steyvers2022bayesian,choudhary2023human}).
This differs from our work in that prior literature assumes oracle ground truth exists (separate from the human experts) and also assumes that the cost of querying experts is negligible.

\paragraph{Online Active Model Selection} Another topic of relevance is online active model selection (OAMS), where pre-trained model predictions from several classifiers are provided at no cost and an algorithm must learn to make sequential predictions on a stream of data \citep{karimi2021online}. Given a fixed budget, these methods can query ground truth at a cost, with the ultimate goal of identifying an optimal policy for routing samples among the provided models. Our work is distinct from OAMS in that we only get predictions from a single classifier and re-define ground truth in terms of human expert consensus instead of an oracle. In our experimental results we adapt the approach of \cite{karimi2021online} to create a baseline for comparison with our proposed methods.

%% file: sections/methodology.tex

\section{METHODOLOGY}
\label{sec:methodology}

\paragraph{Problem Setting} 
Consider a stream of inputs $x_t\in\mathcal{X}\subset \mathbb{R}^{d}$ with dimension $d$ for time $t=1,\dots,T$. We are interested in predicting an associated class $y_t\in\mathcal{Y} := \{1,\dots,K\}$ at each time $t$. Furthermore, we would like to do so in an online fashion by predicting $y_t \sep x_t$ using the information seen in the previous $t-1$ timesteps. We assume unlimited access to a fixed, pre-trained classifier $f(x)$ that produces a probability vector over classes; however, the expected performance and calibration of $f$ is not known beforehand. For brevity, we will refer to specific predictions of $f(x_t)$ as $f_t$.

A key distinction from other online settings is how the ground truth class $y_t$ is determined. We consider the scenario where, given a fixed pool of $N$ human experts, each member votes on the corresponding class of each sample $x_t$. These votes are denoted as $h_t^i \in \{1,\dots,K\}$ for $t=1,\dots,T$ and $i=1,\dots,N$. For convenience, we will denote $H_t^i := \sum_{j=1}^i \texttt{one-hot}(h_t^j) \subset \{0,1,\dots,N\}^K$ as the histogram of the first $i$ votes queried at time $t$. For each timestep, the ground truth is determined as the \emph{consensus} or plurality of human votes, $y_t:=\argmax_k (H_t)_k$ where $H_t := H_t^N$ and $(z)_k$ is the $k^\text{th}$ element of the vector $z$.
In the case of ties, consensus is determined randomly.

Throughout the prediction process, we may either make an immediate prediction based on current beliefs or sample one or more remaining human votes for that timestep. Sampling is conducted one-at-a-time to allow current feedback to further inform the decision making process. We will denote the number of votes drawn so far at time $t$ for input $x_t$ as $N_t \in \{0, 1, \dots, N\}$ and the summary (histogram) of all votes seen as $H_t^{N_t}$. The total running cost is $C_t:=\sum_{t'=1}^t N_{t'}$. An ideal method will produce the lowest achievable error rate when predicting  $y_1, \dots, y_T$ for any fixed average total cost $C_T$.

\paragraph{Probabilistic Model}
We take a probabilistic approach to modeling the generative process of the data to account for uncertainty while querying experts and making predictions. This also confers a convenient means of integrating existing predictions via $f$.

For a given timestep $t\in\{1,\dots,T\}$, there exists a population (effectively infinite) of human opinions concerning the class of $x_t$. By assuming individual votes to be non-fractional  and conditionally independent, we have the following generative distributions:
\begin{align}
\pi_t & \sim \text{Dirichlet}(\alpha_t) \label{eq:inf_pop_dist}\\
H_t & \sim \text{Multinomial}(N, \pi_t) \label{eq:fin_pop_dist}
\end{align}
where $\pi_t$ is the distribution of beliefs about the class of $x_t$ over the population of human experts and $\alpha_t$ represent prior beliefs over the concentration of $\pi_t$. $H_t$ can be thought of as the finite sample of experts available to query at inference time and determines ground truth via $y_t:=\argmax_k (H_t)_k$. When deciding whether to predict or query additional votes, we have access to $N_t$ votes currently drawn as a sub-sample from $H_t$. Since these are drawn without replacement, this sub-sample is distributed as
\begin{align}
H_t^{N_t} & \sim \text{HyperGeo}_K(N, N_t, H_t) \label{eq:fin_subsample}
\end{align}
where $\text{HyperGeo}_K$ is a $K$-dimensional hypergeometric distribution  of $N$ items, $N_t$ separate draws, and $H_t$ contents to subsample \citep{johnson1997discrete}. Due to conjugacy, \cref{eq:inf_pop_dist,eq:fin_pop_dist} exhibit closed form posterior distributions in the presence of the first $i$ votes $H_t^i$:
\begin{align}
\pi_t \sep H_t^i & \sim \text{Dirichlet}(\alpha_t + H_t^i) \label{eq:post_inf_pop_dist}\\
\text{and } H_t^j \sep H_t^i & \sim H_t^i + \text{DirMult}(j-i, \alpha_t + H_t^i) \label{eq:post_fin_pop_dist}
\end{align}
where $j=i+1,\dots,N$ and $\text{DirMult}$ is a compound Dirichlet-Multinomial distribution. Knowing the posterior $H_t^j$ for when $j < N$ is useful for analyzing the potential next $j-i$ queried votes.

\paragraph{A Useful Approximation}
Our primary object of interest is $H_t$, which determines the consensus class $y$. We note that the consensus class does not change if we use the normalized proportion of votes instead of total vote counts:
\begin{align}
y_t:=\argmax_{k\in\mathcal{Y}}(H_t)_k\equiv \argmax_{k\in\mathcal{Y}}\left(\frac{H_t}{N}\right)_k
\end{align}
As such, we note the following limiting case as the finite number of experts $N$ grows:
\begin{align}
\frac{H_t^N}{N} \sep H_t^i & \sim \frac{H_t^i}{N} + \frac{1}{N}\text{DirMult}(N-i, \alpha_t + H_t^i) \nonumber \\
& \overset{d}{\longrightarrow} \pi_t \sep H_t^i \text{ as } N \rightarrow \infty \label{eq:limiting_case}
\end{align}
for some fixed $i\in\mathbb{N}$.
(See the Appendix for a proof). Thus, in the limiting case $y_t:=\argmax_k(\pi_t)_k$ which matches what intuition would tell us. This limiting case can be used as an approximation when $N$ is large, and potentially simplify computations. For brevity, we will refer to the exact, finite expert setting described previously as \fs and this approximate, infinite expert setting as \is. 

\begin{figure}
\begin{center}
    \includegraphics[width=0.8\columnwidth]{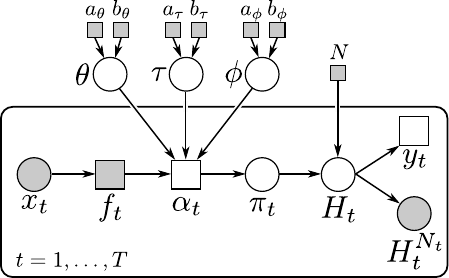}
\caption{Graphical model of the generative process. Grey symbols are observed and assumed known, white are random and unobserved. Circles are random variables and rectangles are deterministic values / transformations. We are interested in estimating $H_t$ for all $x_t$.}
\label{fig:graph_model}
\end{center}
\end{figure}

\paragraph{Incorporating Prior Predictions via $f$}
Determining $\alpha_t$ in \ref{eq:inf_pop_dist}, which controls our prior beliefs over the population-level distribution of classes $\pi_t$, allows  incorporation of the predictions of the model $f_t$ with the human votes. We do so in the following manner with (regularized) positive coefficients: 
\begin{align}
\theta & \sim \text{Gamma}(a_\theta, b_\theta) \notag\\
\phi & \sim \text{Gamma}(a_\phi, b_\phi) \notag\\
(\tau)_k & \sim \text{Gamma}(a_\tau, b_\tau) \text{ for } k=1,\dots,K \notag\\
\alpha_t & := \theta \cdot \text{softmax}(\tau \odot \log f_t) + \phi \label{eq:alpha}
\end{align}
where $\odot$ is the element-wise product, and $a$ and $b$ values are hyperparameters.\footnote{While we restrict attention in this paper to only having access to a single pretrained model $f$, this setup allows for easily extending to multiple models $f^i$ for $i=1,\dots,M$, each with individual model calibration. Simply have $(\tau_i)_k \sim \text{Gamma}(a_\tau, b_\tau)$ and $\alpha_t := \theta \cdot \text{softmax}\left(\sum_{i} \tau_i \odot \log f_t^i\right) + \phi$.} We choose this simple transformation to allow for clear interpretation of parameter values. 
$\phi$ can be thought of as representing the base level of uncertainty concerning all classes as it determines the lower bound on $\alpha_t$. $\theta$ directly quantifies how many expert votes a pretrained model's beliefs are worth. Finally, $(\tau)_k$ enables calibration of $f$ for predictions concerning class $k$. High values of $(\tau)_k$ indicate trustworthy and well-calibrated predictions, and vice versa for low $(\tau)_k$ values. The full set of learnable parameters will be denoted with $\Theta:=(\theta, \phi, \tau)$.  \cref{fig:graph_model} shows the corresponding   graphical model.

\paragraph{Learning Objective and Optimization}
Let $\mathcal{D}_t := \{(f_i, H_i^{N_i})\}_{i=1}^t$ be the collection of model predictions and accumulated votes seen up to time $t$. To perform inference for $t+1$, we must first find the posterior for the learnable parameters: $\Theta \sep \mathcal{D}_t$. This allows for a more informed $\alpha_{t+1}$ which will influence further inference.

While this posterior could be found using Markov-chain Monte-Carlo techniques or approximated with variational inference, %
for computational convenience we instead work with the \emph{maximum a posteriori} (MAP) of $\Theta$.%
Specifically, we compute %
\begin{align}
\Theta^*_t & := \argmin_{\Theta} \mathcal{L}(\Theta; \mathcal{D}_t) \\
\mathcal{L}(\Theta; \mathcal{D}_t) & := \sum_{i=1}^t \log p(H_i^{N_i} \sep f_i, \Theta) + \log p(\Theta)\notag
\end{align}
where $\log p(\Theta) = \log p(\theta) + \log p(\phi) + \sum_i \log(\tau)_i$.
The likelihood of seeing $N_i$ votes changes depending on whether we are using \is or \fs. For the former, the likelihood is:
\begin{align}
& p(H_i^{N_i} \sep f_i, \Theta) \nonumber \\
& := \text{DirMult}(H_i^{N_i}; N_i, \alpha_i) \nonumber \\
& := \frac{\Gamma(\overline{\alpha}_i)\Gamma(N_i+1)}{\Gamma(\overline{\alpha}_i+N_i)}\prod_{k=1}^{K} \frac{\Gamma((\alpha_i + H_i^{N_i})_k)}{\Gamma((\alpha_i)_k)\Gamma((H_i^{N_i})_k+1)}
\end{align}
where $\alpha_i$ is defined in \cref{eq:alpha} and $\overline{\alpha}_i:=\sum_{k=1}^K (\alpha_i)_k$. Unfortunately, for \fs the analytic form of $p(H_i^{N_i} \sep f_i, \Theta)$ involves a sum with $\binom{N-N_i+K-1}{K-1}$ terms, which can be prohibitively large to compute. To avoid this, we represent this likelihood as an expected value to facilitate Monte-Carlo estimation:
\begin{align}
p(H_i^{N_i} \sep f_i, \Theta) & := \E_{H_i \sep \alpha_i} \left[p(H_i^{N_i}\sep H_i)\right] \label{eq:p_h}\\
p(H_i^{N_i}\sep H_i) & := \binom{N}{N_i}^{-1}\prod_{k=1}^K \binom{(H_i)_k}{(H_i^{N_i})_k}.
\end{align}

Unfortunately, the resulting expectation is with respect to a discrete distribution and thus cannot be differentiably sampled from, which is required for gradient-based optimization. To address this, we apply importance sampling with a proposal distribution $q$ that does not rely on the parameters of interest. Leveraging currently seen votes $H_i^{N_i}$ may result in lower estimator variances by biasing samples towards similar proportions of values; however, all possible values of $H_i$ (that do not result in $p(H_i^{N_i}\sep H_i)=0$) must be present in the support of $q$ to ensure valid importance sampling. This leads to the following proposal distribution:
\begin{align}
H_i \sim_q H_i^{N_i} + \text{Multinomial}\left(N-N_i, \frac{H_i^{N_i} + 1}{N_i + K}\right)
\end{align}
where we ensure all classes have support by ensuring the multinomial probability vector has all non-zero values adding 1 to $H_i^{N_i}$ prior to normalizing. Applying importance sampling with this proposal distribution to \cref{eq:p_h} yields the following form:
\begin{align}
p(H_i^{N_i} \sep f_i, \Theta) & = \E^q_{H_i}\!\left[\frac{ p(H_i \sep \alpha_i)}{q(H_i\sep H_i^{N_i})}p(H_i^{N_i}\sep H_i)\right],
\end{align}
where $p(H_i \sep \alpha_i) := \text{DirMult}(H_i; N, \alpha_i)$ and $\E^q$ is the expectation with respect to the  proposal distribution $q$.

Thus, with $M$ Monte-Carlo samples, the likelihood can be approximated via
\begin{align}
p(H_i^{N_i} \sep f_i, \Theta) & \approx \frac{1}{M} \sum_{j=1}^M \frac{p(H_i^{(j)} \sep \alpha_i)}{q(H_i^{(j)}\sep H_i^{N_i})} p(H_i^{N_i}\sep H_i^{(j)}) \notag
\end{align}
where $H_i^{(j)} \sim q(H_i \sep H_i^{N_i})$ for $j=1,\dots,M$. Computing this for $i=1,\dots,t$ allows for the computation of $\mathcal{L}(\Theta; \mathcal{D}_t)$ in a differentiable manner.

\paragraph{Decision Making}
Assume that the MAP estimate for $\Theta \sep \mathcal{D}_{t-1}$ has been found, resulting in $\alpha_t^* = \theta^* \cdot \text{softmax}(\tau^* \odot \log f_t) + \phi^*$. For generality, also assume we have already seen $N_t > 0$ votes for the current time step $t$, and are deciding whether to continue querying new votes or to make a prediction for $y_t$. 

The model has beliefs over the true consensus class $y_t for$ $x_t$. In \fs, it follows that:
\begin{align}
p(y_t=k & \sep  H_t^{N_t}, f_t,  \mathcal{D}_{t-1}) \approx p(y_t=k \sep H_t^{N_t}, \alpha_t^*) \nonumber \\
& = p({\argmax}_{k'} (H_t)_{k'} = k \sep H_t^{N_t}, \alpha_t^*) \nonumber \\
& = \E_{H_t \sep H_t^{N_t}, \alpha_t^*} \left[\ind({\argmax}_{k'} (H_t)_{k'} = k)\right]
\end{align}
where $\ind(\cdot)$ is the indicator function, returning 1 if the argument is true and 0 if false.\footnote{In the case of a simulated $H_t$ resulting in a tie, we randomly choose one of the tied classes to be the $\arg\max$, 
in the same manner as ground truth is determined.} This expected value can be approximated using a Monte-Carlo estimate. For \is, the formula is the same, aside from swapping $H_t$ for $\pi_t$. This is justified using the same reasoning as to the existence of the limiting case in the first place, as demonstrated in \cref{eq:limiting_case}.

Given current information, choosing the $\hat{y}_t$ with the highest likelihood, i.e., $\hat{y}_t := \argmax_k p(y_t=k \sep  H_t^{N_t}, \alpha_t^*)$ represents the optimal decision under the posterior. Additionally, assuming our model is well-calibrated, $p(y_t=\hat{y}_t  \sep  H_t^{N_t}, \alpha_t^*)$ can be interpreted as the expected accuracy of our prediction. This probability will be denoted as $\acc_t(H_t^{N_t})$. One obvious choice of heuristic for determining when to stop querying is to simply predict once the accuracy has cleared some set threshold: $\acc_t(H_t^{N_t}) > \rho$. We utilize this threshold decision on estimated accuracy as our heuristic for experiments detailed in \cref{sec:experiments_results}.

%% file: sections/experiments.tex

\section{EXPERIMENTS AND RESULTS}
\label{sec:experiments_results}

\begin{figure}[t]
\includegraphics[width=0.985\columnwidth]{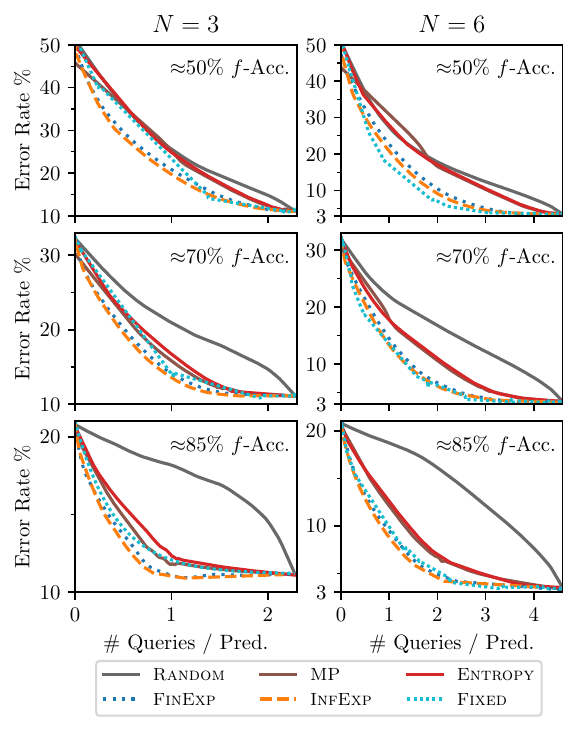}
\caption{Error-cost curves similar to \cref{fig:err_cost_c10h} except on ImageNet-16H with $N=3,6$.}
\label{fig:err_cost_i16h}
\end{figure}

We evaluate our online consensus prediction methods on two large-scale datasets that include multiple per-sample human predictions. For each, samples $x_t$ are drawn randomly from the test set without replacement to create a sequence. At a given timestep $t$, each method is given predicted model confidences $f_t$ for sample $x_t$ and the opportunity to query expert votes $h_t$ or make a final prediction. Expert votes can be queried sequentially, allowing methods to re-evaluate under new information. Once the method ceases querying, it then generates a final prediction and proceeds to the next timestep. As noted in \cref{sec:methodology}, we assume a cost of $1$ for querying a human expert. In addition, we utilize several different pre-trained neural network models $f$ with varying performances to evaluate the robustness of our findings. Below, \cref{fig:err_cost_c10h,fig:err_cost_i16h} illustrate error rate as a function of querying budget, standardized over sequence length as the average number of queries taken per sample. The y-axis origin of these plots represents the empirical lower bound on error rate due to ties in consensus. Overall, the experiments demonstrate improved predictive performance for the proposed methods compared to baselines. Additionally, our proposed method exhibits robustness to distribution shift. All experiments were conducted on NVIDIA GeForce 2080ti GPUs over roughly 4 days. More experimental details can be found in the Appendix. 

\subsection{Datasets}

For our experiments we utilize CIFAR-10H \citep{peterson2019human} (an extension of the CIFAR-10 \citep{krizhevsky2009learning} test dataset that contains 10,000 natural images categorized into 10 classes). Each image in CIFAR-10H also includes 50 human predictions, allowing us to explore multiple expert pool sizes by first randomly selecting $N \leq 50$ experts and then generating randomly ordered sequences of length 10,000 from the test set. We present results for $N=3$ and $N=50$, with additional results in the Appendix. Consensus is determined from this sampled expert pool, with ties broken randomly. We also train several ResNet18 models \citep{he2016deep} to utilize as $f$, using subsets of the CIFAR-10 training dataset with varying levels of class imbalance, yielding models with significantly different error rates. 

We additionally leverage the ImageNet-16H dataset \citep{steyvers2022bayesian}, which contains 4800 natural images grouped into 16 classes with 4 different levels of phase noise  (none, low, medium, and high noise, 1200 images each). \cite{steyvers2022bayesian} also included 5 different classifiers pre-trained on the ImageNet-16H training data with varying error rates. Four variants of each classifier were then trained for 0, 0.5, 1, or 10 epochs on noised training data, increasingly improving model performance. In this dataset, each image is annotated by 6 experts. Similar to CIFAR-10H, we present results for $N=3$ and $N=6$, and we define $y$ similarly as the consensus of available experts.

\subsection{Baselines and Methods}

We compare against several baselines motivated by the active learning and online prediction literature and adapt these methods to accommodate online consensus prediction.  
As a simple baseline, we utilize a random (\textsc{Random}) predictor. This method is parametrized by a single global Bernoulli parameter $\beta$ and ignores model predictions $f_t$ when querying. For each example $x_t$, we draw $\text{Binom}(N, \beta)$ number of expert votes and predict the $\arg\max$ of them. Should no experts be queried, then $\arg\max_k (f_t)_k$ is used as a prediction. The other two baselines both use model predictions to influence $\beta$ on a per-example basis. The \textsc{Entropy} baseline uses the prediction entropy $H(f_t)$ of the model's predictions, scaled and restricted to the range $[0,1]$, to define $\beta_t$, which is then used in the same manner as the random baseline. The third baseline is based on the Model Picker method from \cite{karimi2021online} (\textsc{MP}), which adapts the Exponential Weights algorithm \cite{littlestone1994weighted} to dynamically maintain a loss-estimate over several models. Since we utilize a single model in our evaluation, we tailor \textsc{MP} to track per-class loss estimates. The method uses approximate loss estimates over classes plus the model's label distribution for sample $x_t$ to aggregate into a single Bernoulli parameter $\beta_t$. Queries are then determined in the same manner as the \textsc{Entropy} and \textsc{Random} baselines. Additional details can be found in the Appendix.

In the context of our proposed framework, we apply \fs and \is in the following manner. We learn a MAP estimate of $\Theta$ and leverage it to create $\alpha^{*}_t$ via \cref{eq:alpha}. For each sample $x_t$, querying is decided by thresholding $\max p(y_t=k  \sep  H_t^{N_t}, \alpha_t^*)$ with hyperparameter $\rho$. If an expert is queried, we then update our observed expert votes $H_t^{N_t}$ and repeat the process. Otherwise, $\hat{y} := \argmax_k p(y_t=k  \sep  H_t^{N_t}, \alpha_t^*)$ is submitted as the final prediction.

\begin{figure}[t]
\includegraphics[width=0.99\columnwidth]{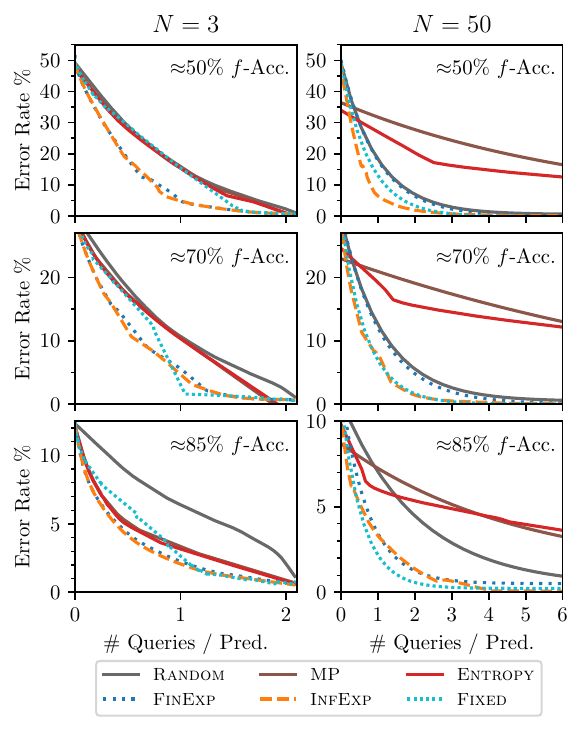}
\caption{Error-cost curves on CIFAR-10H with $N=3,50$ and model accuracies of $50\%, 70\%$, and $85\%$. Solid and dotted curves are baselines and proposed methods.}%
\label{fig:err_cost_c10h}
\end{figure}

In addition to \fs and \is, we seek to isolate the impact of our learned MAP estimates from the inference procedure. In turn, we fix the prior parameters $\theta = \phi = \tau = 1$ and do not update them during inference. We will refer to this method as \fx. For space, we only present results for \fx under the \fs inference procedure and refer the reader to the Appendix for the \is equivalent. Finally, we note several computational speedups that yield a negligible performance difference with the naive implementation. We utilize the latter in all experiments and note a 20x speedup. Please see the Appendix for further details.

\subsection{Results: Error-Cost Curves}
Using the methodology described above, we generate error-cost curves for each of our proposed methods and baselines, for each dataset, and combined with models of different accuracies.
Given a particular randomly-ordered sequence from the test set, each method sequentially handles querying and prediction for each example $x_t$ 
and the overall per-sample querying cost and error rate are then plotted as a single point. By sweeping over hyper-parameter settings and 3 trials, we generate an error-cost curve that reflects expected error rates for a range of budgets.
Please see the Appendix for additional implementation details. 

Figures \ref{fig:err_cost_c10h} and \ref{fig:err_cost_i16h} show the   cost-error results for CIFAR-10H and ImageNet-16H, respectively. We evaluate all methods on different sizes of $N$ using three difference models with approximately 50\%, 70\%, and 85\% accuracy relative to consensus ground truth. For ImageNet-16H we see that the \fs and \is methods consistently outperform all baselines across all budget settings with all models. 
In some cases, \fx also outperforms the baselines, even achieving the outright best performance with $N=6$ and base model accuracy of 50\%. However, the performance of \fx is generally inconsistent across settings.

\begin{figure}
\includegraphics[width=0.93\columnwidth]{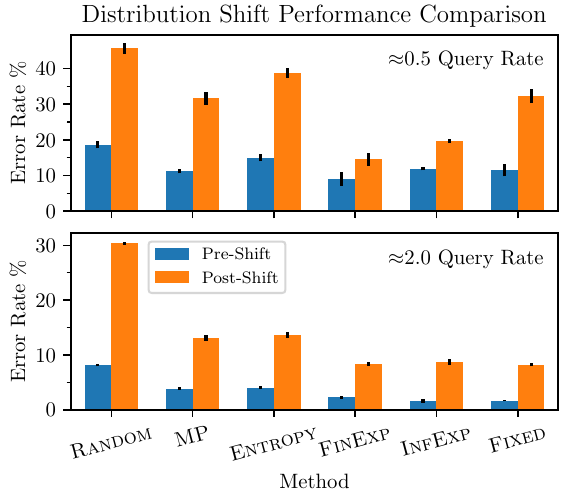}
\caption{Visualization of aggregate distribution shift error rate on ImageNet-16H with a DenseNet model $f$ for budgets of 0.5 and 2 queries per sample.}
\label{fig:dist_shift_comparison}
\end{figure}

\begin{figure}[t]
\includegraphics[width=0.99\columnwidth]{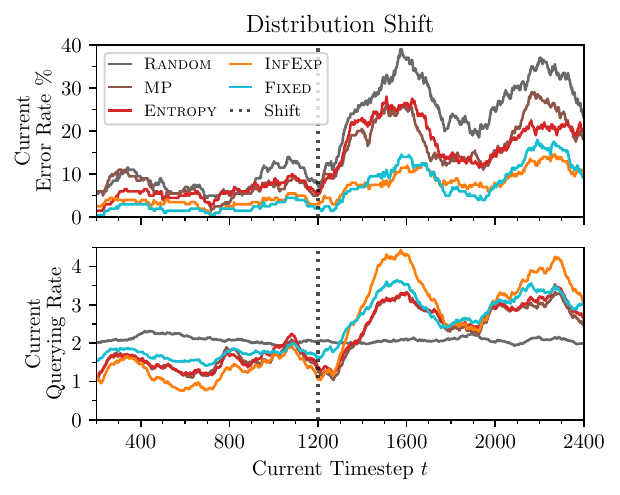}
  \caption{Visualization of distribution shift error rate and querying cost on ImageNet-16H with a DenseNet model $f$ and an average querying cost of 2 experts per sample. Distribution shift occurs at $t=1200$, increasing uncertainty in model predictions and expert consensus.}
\label{fig:dist_shift}
\end{figure}

These trends continue when analyzing results from CIFAR-10H in \cref{fig:err_cost_c10h} with dramatic improvements observed for \is over all settings. \fs also outperforms all baselines, though by a very small margin over the Random baseline for certain settings where $N = 50$. In general we found \is to be more consistent than \fs across different settings. We conjecture these results as a whole to be attributable to both the online calibration of the model predictions through parameters $\tau$ for \fs and \is, and also to tying our decision to query (based on the model's beliefs) directly to expert consensus through $\hat{y}_t := \arg\max_k (H_t)_k$ (or $(\pi_t)_k$ for \is).

\subsection{Results: Out-of-Distribution Performance}

To further explore the robustness of our methods under distribution shift, we refined the ImageNet-16H dataset such that each test sequence consists of  1200 non-noisy images (in random order)  followed by a random sequence of 1200 high-noise images. We conduct online consensus prediction as before for our cost-error models, using the five pre-trained models that never observed noisy data. This in effect induces a distribution shift  in the test sequences after 1200 samples, causing models to suddenly encounter much  noisier images than those on which it was trained.

To compare our results, we run all methods and baselines across a hyperparameter sweep and multiple trials. Then, we group all runs that fall within 10\% of 0.5, 1, 2, and 3 average querying costs to produce aggregate results. Shown in \cref{fig:dist_shift_comparison}, our proposed methods possess lower error rates and demonstrate significantly lower performance degradation after the distribution shift occurs, even when pre-shift error rates are comparable.

\cref{fig:dist_shift} more granularly explores the distribution-shift behavior of all methods by focusing only on a subset of runs that all possess an average cost of 2 queries per sample. Both before and after distribution shift, \is and \fx consistently outperform the baselines. Interestingly, though \is and \fx maintain nearly identical error rates, \is has the \textit{lowest} pre-shift and the \textit{highest} post-shift querying costs, demonstrating the flexibility conferred by dynamically learning a prior under distribution shift.  This is also highlighted by visualizing the $\tau$ parameters against model performance, as shown in \cref{fig:tau_analysis}. To avoid clutter, we only plot the results of \is and \fx, with further results provided in the Appendix. Due to stochasticity in the sequence ordering as well as the effect of the moving average, we note fluctuations performance over time for all methods. We explore these fluctuations for additional properties like distribution shift convergence in the Appendix by averaging over many sequence orderings for the same budget.

\begin{table}
\centering
\fontsize{8pt}{8pt}\selectfont
\caption{Query-free accuracy improvement over base model $f$ under different model accuracy and querying budget settings for CIFAR-10H.}%
\label{tab:two_phase}
\begin{tabular}{clccc}
\toprule
\multicolumn{2}{c}{Queries / Sample:} \hspace{-1em} &  \multicolumn{1}{c}{1} & \multicolumn{1}{c}{2} & \multicolumn{1}{c}{3}\\ 
\midrule
  Acc.\hspace{-1em} & Method \hspace{-1em} & \multicolumn{3}{c}{Acc. Improvement Over $f$}
 \\
\midrule
\multirow{4}*{50\%}\hspace{-1em} & \is \hspace{-1em}  & 0.99±0.08     & 1.01±0.09     & 1.09±0.14    \\ 
 & \xis \hspace{-1em}  & 1.21±0.11     & 1.10±0.06     & 0.98±0.15    \\ 
 & \fs \hspace{-1em}  & 1.45±0.17     & 1.33±0.13     & 1.98±0.22    \\ 
 & \xfs \hspace{-1em}  & \textbf{1.59±0.34}     & \textbf{2.79±0.08}     & \textbf{3.29±0.23}    \\ 
\midrule
\multirow{4}*{70\%}\hspace{-1em} & \is \hspace{-1em}  & 0.77±0.08     & 0.63±0.04     & 0.59±0.11    \\ 
 & \xis \hspace{-1em}  & 0.81±0.06     & 0.69±0.03     & 0.5±0.03    \\ 
 & \fs \hspace{-1em}  & \textbf{1.05±0.08}     & 0.83±0.08     & 1.08±0.06    \\ 
 & \xfs \hspace{-1em}  & 0.90±0.11     & \textbf{1.07±0.03}     & \textbf{1.24±0.07}    \\ 
\midrule
\multirow{4}*{85\%}\hspace{-1em} & \is \hspace{-1em}  & 0.13±0.02     & 0.07±0.01     & 0.07±0.02    \\ 
 & \xis \hspace{-1em}  & 0.11±0.01     & 0.09±0.00     & 0.13±0.02     \\ 
 & \fs \hspace{-1em}  & 0.15±0.04     & \textbf{0.16±0.06}     & \textbf{0.32±0.03}    \\ 
 & \xfs \hspace{-1em}  & \textbf{0.15±0.02}     & 0.15±0.01     & 0.24±0.04    \\   
  \bottomrule                          
\end{tabular}
\end{table}

\subsection{Results: Calibration and Query-Free Prediction} 

In a practical setting, there may be situations where human experts will be unavailable, such as nights, weekends, holidays, or periods where everyone is busy. We explore these scenarios with a suite of two-phase experiments where we first query and predict as normal but then enter a second phase where no human experts are available to query and we must solely rely on the learned parameters and pretrained model predictions. We use both   \is and \fs as described for both querying and predicting. We also consider the scenario of using the \fx protocol for querying while learning and predicting with \is and \fs. We refer to these options as \fx-\is and \fx-\fs respectively. Results can be seen in \cref{tab:two_phase}.
We find a small but consistent average performance increase of  up to 3\% over the base model accuracy in many settings, but particularly in cases where the querying budget is large and base model performance is low. This demonstrates the ability for the MAP parameter estimates to learn per-class model performance through noisy expert votes and calibrate the pretrained model's predictions. 

\begin{figure}[t]
\includegraphics[width=0.99\columnwidth]{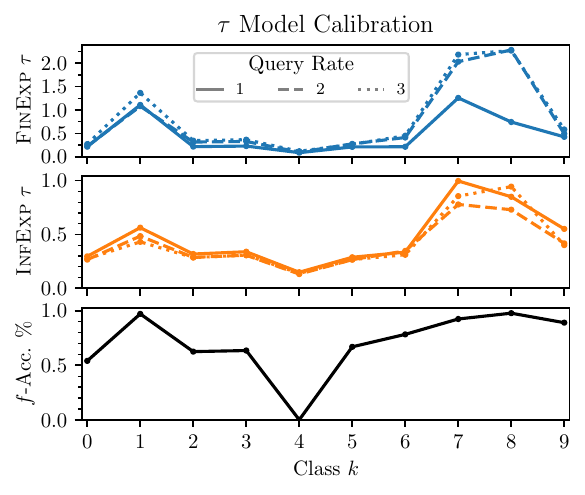}
  \caption{Visualization of per-class performance of a CIFAR-10 model with 70\% accuracy under  learned $\tau$ parameters from \fs and \is with different querying budget rates.}%
\label{fig:tau_analysis}
\end{figure}

To visualize this capability, we plot the performance of a pretrained model with extremely varied per-class accuracy on CIFAR-10H against the learned $\tau$ values of \fs and \is, shown in \cref{fig:tau_analysis}. In addition to correlating strongly with the true predictions, we observe that \fs more heavily weights its beliefs with new evidence, while \is, which assumes a pool of experts, does not. This can be seen by noting the increased values in the blue curves from solid to dotted/dashed (especially for classes 7, 8, and 9), while the different orange curves remain all roughly the same. Please see the Appendix for a more thorough analysis.

%% file: sections/discussion.tex

\section{CONCLUSION}

\paragraph{Limitations and Future Directions} 

Our experimental results rely only on two crowdsourced datasets from a single data modality (images). %
Nonetheless,  the strength of our empirical findings 
offer a robust starting point for future work in online consensus prediction. 
 
The work in this paper leaves multiple  future directions to explore. For example, we did not consider adversarial data streams, non-stationary streams, or open-set tasks where new classes may be progressively observed. Exploring these scenarios is a promising opportunity for future work. Further, allowing experts to be identifiable, ranked (e.g. by seniority), and variable in cost and/or availability offers new dimensions to consensus prediction.

\paragraph{Conclusion}
In this paper we  proposed the novel and practical problem of online human consensus estimation, where the target $y$ is  defined as the consensus of a pool of human experts. We approached this problem in a Bayesian fashion, offering a principled means of reasoning about human expert consensus given a subset of votes and a pre-trained model $f$ with unknown performance.
We derived \fs, based on the multivariate hypergeometric likelihood, and also introduced the \is method, a computationally simple limiting case of \fs when $N\rightarrow\infty$. Empirically, we evaluated \fs and \is against several baselines, including \fx, a variant of \fs with a fixed prior. Our results demonstrate the promise of \fs and \is both in standard experimental settings as well as with distribution shift.

%% file: sections/acknowledgements.tex
\subsection*{Acknowledgements:}
This work
was supported by National Science Foundation Graduate
Research Fellowship grant DGE-1839285 
(SS and AB), by the National Science Foundation under award number 1900644 (PS and MS),
by the National Institute of Health under awards R01-AG065330-02S1 and R01-LM013344 (PS), by the HPI Research Center in Machine Learning and Data Science at
UC Irvine (SS and PS), and by Qualcomm Faculty awards (PS).

%% file: appendix/limiting_case.tex

\section{Proof for \textsc{InfExp} being a Limiting Case of \textsc{FinExp} (Section 3 in paper)} %
\label{sec:Proof for Limiting Case of FinSet}

Below, we outline and prove that the limiting case of \fs is \is as $N \rightarrow \infty$.

For $\theta \sim \text{Dirichlet}(\alpha)$ with $K$ dimensions and $h_i \sep \pi \sim \text{Multinomial}(1, \pi)$ for $i=1,2,\dots$, let $H^n := \sum_{i=1}^n \texttt{one-hot}(h_i)$ for $n > 0$ and $H^0=\vec{0}$. It follows that $H^n \sep \pi \sim \text{Multinomial}(n, \pi)$, and that $\theta \sep H^n \sim \text{Dirichlet}(\alpha + H^n)$. $H^n$ also is known to have an unconditional distribution of $H^n \sim \text{DirMult}(n, \alpha)$ and posterior $H^N \sep H^n \sim H^n + \text{DirMult}(N-n, \alpha)$ for $N > n$. Note that this is shorthand for denoting $H^N - H^n \sep H^n \sim \text{DirMult}(N-n, \alpha)$.
\\\\
The posterior distribution of the normalized counts, $H^N / N$, converges in distribution to the posterior of $\pi$. Put formally:
\begin{align}
\frac{H^N}{N} \sep H^n & \sim \frac{H^n}{N} + \frac{1}{N}\text{DirMult}(N-n, \alpha + H^n) \nonumber \\
& \overset{d}{\longrightarrow} \pi \sep H^n \text{ as } N \rightarrow \infty \end{align}
for some fixed $n\in\mathbb{N}$

\textit{Proof:} First, we note that we are interested in the limiting case of the random variable $\frac{H^N}{N} - \frac{H^n}{N} \sep H^n$. Since we condition on $H^n$, $\frac{H^n}{N}$ can be treated as a constant term with respect to $N$, which converges to 0 as $N\rightarrow \infty$. Thus, analyzing the limiting case of $\frac{H^N}{N} \sep H^n$ is sufficient.
\\\\
Next, we denote the joint distribution of $H^N, \pi \sep H^n$ as $p(H^n, \pi \sep H^n) = p(\pi \sep H^n)p(H^N \sep \pi) := \text{Dir}(\pi; \alpha + H^n)\text{Mult}(H^N; \pi)$. The distribution of $\pi \sep H^n$ can be treated as a constant with respect to $N$. Lastly, $\frac{H^N}{N} \sep \pi = \frac{1}{N}\sum_{i=1}^N h_i \sep \pi \overset{d}{\longrightarrow} \pi \sep \pi$ as $N \rightarrow \infty$ due to the law of large numbers. Thus we can conclude that $\frac{H^N}{N}, \pi \sep H^n \overset{d}{\longrightarrow} \pi \sep H^n$ as $N\rightarrow\infty$ which implies that $\frac{H^N}{N}\sep H^n \overset{d}{\longrightarrow} \pi\sep H^n$.

%% file: appendix/experimental_setup.tex

\section{Experimental Details (Section 4 in paper)} %
\label{sec:Experimental Setup Details}

\subsection{Datasets} %
\label{sub:Datasets}

\paragraph{CIFAR-10H} %
\label{par:CIFAR-10H}

The CIFAR-10H dataset \cite{peterson2019human} consists of 10,000 samples corresponding to the test-set of CIFAR-10 \cite{krizhevsky2009learning}, with 50 human annotations per label. On average, each individual   human annotator only labeled a few hundred of the test samples: as a result, 50 "full" annotators were synthetically generated by combining annotations from multiple human annotators. 

Model predictions utilized for CIFAR-10H experiments were generated by training ResNet-18 models on random sub-slices of the training dataset with stochastic gradient descent and a learning rate of 0.01. To get a variety of models, checkpoints were stored every 10 epochs and leveraged to provide models with different levels of test-set accuracy. Predictions for these models are included in the code repository noted in the main paper.

\paragraph{ImageNet-16H} %
\label{par:ImageNet-16H}

The ImageNet-16H dataset used in our experiments includes model predictions provided with the original ImageNet-16H paper \cite{steyvers2022bayesian}. These predictions were generated by selecting 5 pre-trained models (AlexNet, DenseNet, VGG-16, ResNet, and GoogLeNet) on the original ImageNet dataset and renomalizing the prediction probabilities over 16 classes. 1200 images belonging to these 16 classes (300 each) were stored and then perturbed with 3 different levels of phase noise, also preserving the original. In total, this created 4800 images. For each image, 6 human annotations are available. Each of the 5 models is then fine-tuned for 0, 0.5, 1, and 10 epochs on noised equivalents of the training dataset, totaling 20 models. Each model produced a prediction for each image. The dataset provided in the repository noted in the main paper includes model predictions to allow for easy experiment replication.

\subsection{Baselines} %
\label{sub:Methods}

\paragraph{Random} %
\label{par:Random}

The \textsc{RANDOM} baseline determines how many experts to query per example as a function of the hyperparameter $\beta$. Given a fixed pool of $N$ experts, the random baseline queries $Q \sim \text{Binom}(N,\beta)$ experts one-at-a-time, stopping early if consensus is reached.

\paragraph{Entropy} %
\label{par:Random}
The \textsc{ENTROPY} baseline first computes the prediction entropy from a pre-trained model $f$ as $\mathcal{H}(f_t) := -\frac{1}{K}\sum^{K}_{i=1}(f_t)_i \log (f_t)_i$. A tuning hyperparameter $v \in \mathbb{R}_{+}$ is then applied and the per-sample query parameter $\beta_t$ is determined as $\beta_t := \max(\min(H(f_t),1),0)$, i.e., the entropy is clipped to the $[0,1]$ range. The number of queries is then determined in the same manner as for the \textsc{RANDOM} baseline.

\paragraph{Model Picker} %
\label{par:Random}

As noted in \cite{karimi2021online}, Model Picker (\textsc{MP}) is a context-free online-active model selection algorithm. We adapt \textsc{MP} to the case of a single model by determining variance calculations over the classes and updating per-class belief accordingly. All other features of MP are left unchanged, except that instead of sampling a single ground truth oracle, \textsc{MP} now samples the number of queried experts from a binomial in the same fashion as the \textsc{RANDOM} and \textsc{ENTROPY} baselines.

\paragraph{Learning MAP Parameter Estimates} %
\label{par:learning_maps}

In our experiments we utilize Gamma distributions $\Gamma(a,b)$ as (hyper-)priors over our learnable parameters $\Theta :=  \{\theta ,\phi , \tau \}$. All learnable parameters utilize the same (relatively flat) prior distribution and are initialized to the mode of the prior before optimization begins. We compute MAP estimates by use of stochastic gradient descent and the Adam optimizer with a learning rate of 0.1. For a given timestep $t$, we define the dataset on which we learn our prior as $\mathcal{D} :=  \{f_{t'}, H_{t'}^{N_{t'}} | t' < t, N_{t'} > 0\}$. In words, we consider all observed data from previous samples, but filter out samples for which we did not query a single expert, as these samples do not contribute to learning MAP parameters. Furthermore, for computational feasibility over long sequences, we further reduce $\mathcal{D}$ to only the last $w$ observed samples, creating a sliding window on which we learn MAP estimates. For our experiments, $w$ is set to 500. In addition, rather than learn MAP estimates for each new sample, we learn them at a fixed interval of 20 iterations for all experiments, noting a negligible difference in converged values. Training is conducted for 1000 iterations or until the maximum difference in the parameter values of $\Theta$ is less than a tolerance of 0.01 between updates across 10 iterations. In general, we note that, with the exception of distribution shift experiments, after a few hundred iterations the values of $\Theta$ appear to converge and experiments accelerate. 

For \fs experiments, a finite number of experts is assumed in the likelihood, regularizing the method more so that the \is counterpart, which assumes an expert pool of infinite size. To reflect this, we utilize priors of $\Gamma(1.1,1)$ and $\Gamma(3,2)$, for \fs and \is, respectively. This is to more heavily constrain the \is prior parameters under the potential for infinite feedback. These settings were applied to all experiments. 

\subsection{Experiment: Error-Cost Curves} %
\label{sub:Error Cost Curves}

In addition to the discussion of error-cost curves provided in the main paper, we include the following additional details to better inform experiment replication. First, a dataset is loaded and a subset of experts is selected from the population pool. A sequence of samples is then drawn from the dataset without replacement and placed in random order. Ground truth is subsequently defined with regard to all experts in the subset, with ties broken randomly. We adjust the origin of the y-axis of the plots in both the main paper and the appendix to reflect the empirical lower bound on error rate due to ties. This is particularly noticeable in ImageNet-16H.  We utilize random seeds 3,4, and 5 to seed all randomness in our experiments. In addition, we evaluate a range of hyperparameter settings for each method to generate full power-cost curves. 

Once ground truth is computed, we begin experiments by sweeping across hyperparameters for each method. The base ranges for hyperparameters is listed below:

\begin{itemize}[noitemsep]
  \item \fs, \is, \fx  - $[0,1]$
  \item \textsc{RANDOM} - $[0,1]$
  \item \textsc{ENTROPY} - $[0,1000]$
  \item \textsc{MP} - $[0,1000]$
\end{itemize}

All of these hyperparameter values are multiplied by the original $\beta$ coefficients for binomial sampling and then clamped to be within the 0-1 range to create a final $\beta$ parameter for sampling. Both the \textsc{ENTROPY} and \textsc{MP} baselines may possess very small values $\beta_t$. To ensure the final $\beta$ parameter fully spans the 0-1 range (and therefore all querying budgets), we sweep across the larger range of 0 to 1000 and clamp values to 1 in cases where the scaled value exceeds this range.

For each error-cost curve in the main paper in \cref{fig:err_cost_c10h} and \cref{fig:err_cost_i16h}, we fit a Lowess smoother with a weight fraction of 0.20 to the resulting scatter plots to visualize a smoothed error-cost curve for all querying budgets.  For each plot, we set the domain of the error-cost curves such that methods achieve the empirical lower bound on performance. That bound is 10.03\% for ImageNet-16H and $N=3$ experts, 2.92\% for the same dataset and $N=6$ experts. The error lower bound for CIFAR-10H, a much less noisy dataset, is less than 0.50\% for all settings of $N$.

\subsection{Experiment: Distribution Shift} %
\label{sub:Distribution Shift}

Distribution shift experiments combine the 1200 clean (un-noised) samples from ImageNet-16H with most-noised samples. All methods are then run on this sequence, which randomly orders the first and second 1200 samples but keeps all of the clean samples in the first half of the experiment to make the distribution shift abrupt. We plot all results, shown in \cref{fig:dist_shift} in the main paper and additional results in \cref{fig:dist_shift_arch,fig:dist_shift_budget}, by utilizing a simple moving average to produce running estimates of error rate and querying cost. Aggregate data explores these trends further in \Cref{fig:dist_shift_comparison}, grouping runs that achieve within 0.1 queries/sample of a budget of 0.5 and 2 experts per sample. Evaluation of additional budgets can be found in the additional experiments below.

\subsection{Experiment: Two-Phase Evaluation} %
\label{sub:two_phase_eval}

In two-phase evaluation experiments, we take the first 1000 data points from a sequence generated from either the CIFAR-10H or ImageNet-16H dataset. Given our method of choice, we query experts as needed for these first 1000 samples. At this point, access to expert feedback stops and methods must predict using just the base model predictions and the MAP parameters of $\Theta$ that they learned on the previously observed data. In this case, we do not utilize a sliding window and instead learn a prior over all of the observed data. We then group experiments that are within 0.1 queries/sample of querying budgets of 1, 2, and 3 experts per sample, producing the mean and standard error or the runs. Our results demonstrate that our proposed methods, \fs, \is, and the \fx method combined with either the \fs or \is inference method (known as \xfs and \xis), all consistently (but slightly) outperform the base model accuracy. These metrics are all relative to the base model accuracy.

\subsection{Experiment: Model Calibration} %
\label{sub:model_cal}

Exploring the results of the two-phase evaluation further, we select a model from CIFAR-10H with highly varied per-class performance. Several different experiments were conducted using this model, and reported in the main paper is a non-cherry-picked set of results for $\tau$ under budgets of 1, 2, and 3 experts per sample. These are plotted against the base model accuracy. In the additional experiments below, we include the aggregate correlations between the learned MAP parameters of \fs and \is and display these results in the following section.

%% file: appendix/additional_experiments.tex

\section{Additional Experimental Results (Sections 4.3 - 4.5 in paper)} %
\label{sec:Additional Experimental Results}

Below, we produce and examine experimental results not reported in the main paper. While the general findings are consistent, we note additional details and insights that extend on our initial results.

\paragraph{Results: Error-Cost Curves (Section 4.3 in paper)} %
\label{par:ErrorCost Curves}

\begin{figure}[h!]
\includegraphics[width=0.99\textwidth]{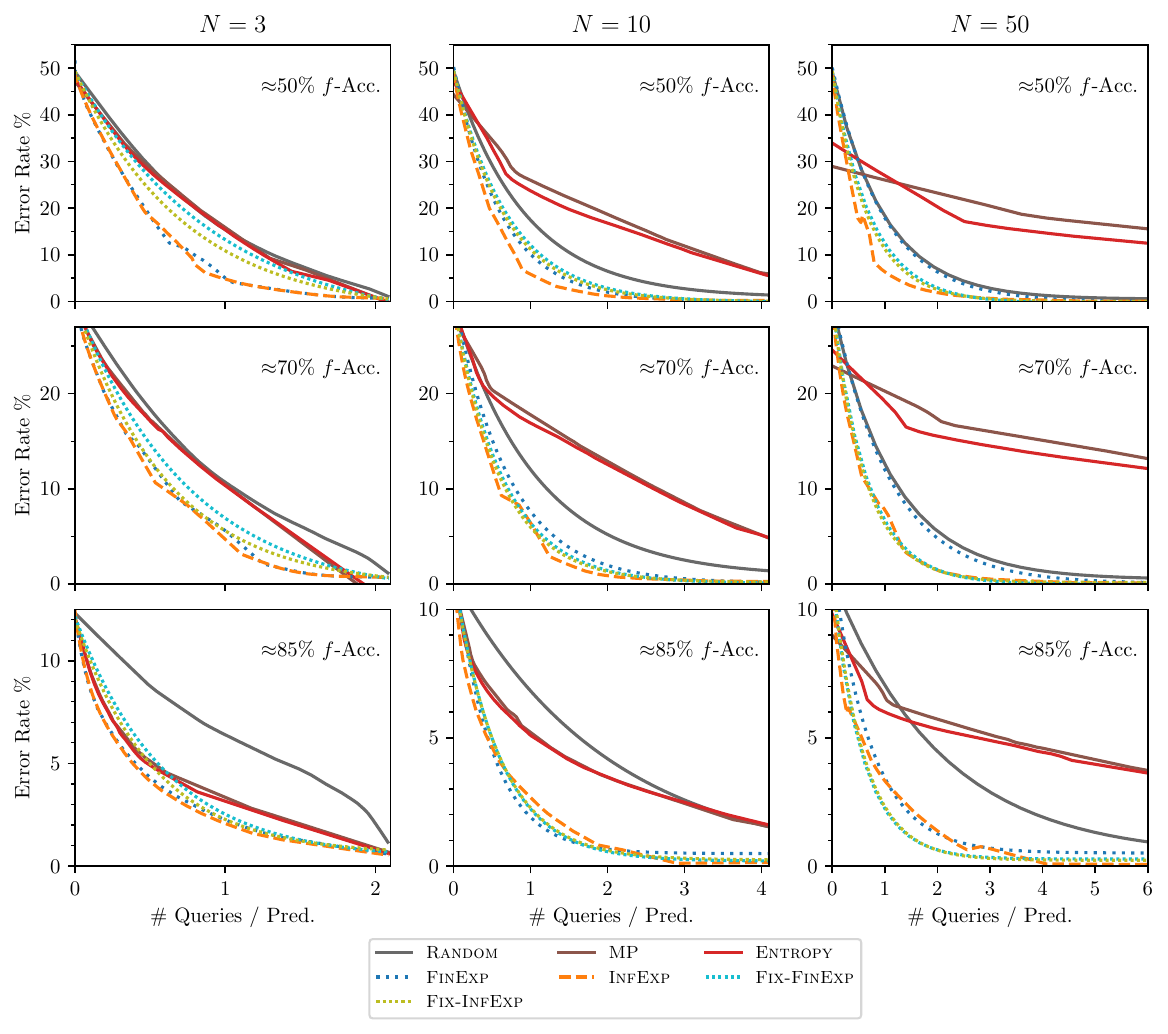}
  \caption{Error-cost plots for CIFAR-10H data with $N=3,10,50$. Fixed baselines \xis and \xfs are also included and show strong but inconsistent results. By contrast, \is and \fs maintain consistent results across a variety of budget and hyperparameter settings.}
\label{fig:appendix_cifar_power_curve}
\end{figure}

In addition to creating error-cost curves for experiments on CIFAR-10H with $N=3,50$, we also include findings for $N=10$. Seen in \Cref{fig:appendix_cifar_power_curve}, we witness similar trends to $N=50$. However, we observe no degradation in performance in \fs across runs, whereas with $N=50$ \fs approaches the random baseline for model accuracies of 50\% and 70\%. 

At the same time, we include the results for the \xis baselines, shown in all experimental settings but omitted from the main paper due to its comparable performance to \fx, which we refer to here as \xfs here for clarity. These represent baselines where no prior parameters are learned but either the \is or \fs procedure is still utilized. In general, our findings remain consistently superior to the baselines, with \is offering the most consistent performance overall. 

Transitioning to additional findings for ImageNet-16H, we include the \xis baseline and observe its high but variable performance. Though superior in many cases, we witness the results of both \xis and \xfs varying significantly across querying budgets as well as experimental settings. The process of learning a prior with either  \fs or \is demonstrably stabilizes empirical performance across hyperparameter settings. In all settings we notice a tendency for \fs to underperform relative to \is and posit these effects to be a result of approximating the likelihood via sampling, whereas the likelihood of \is is available in closed form and does not require sampling.

\begin{figure}[h!]
\includegraphics[width=0.90\textwidth]{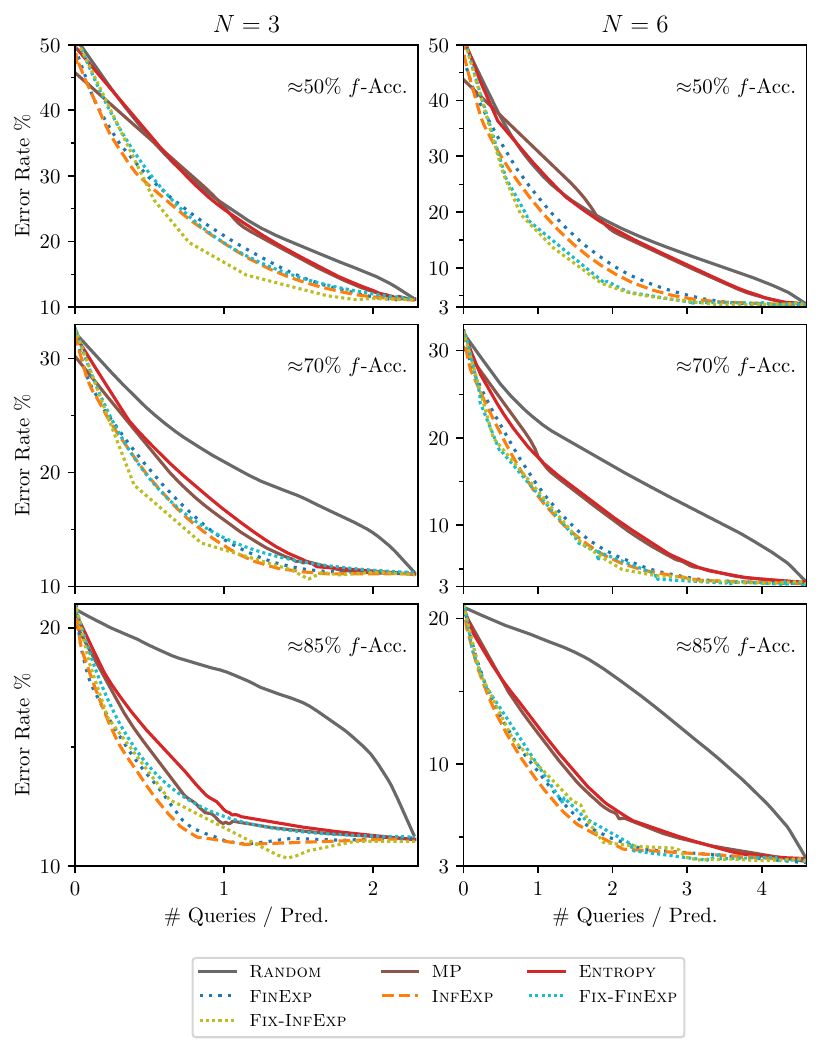}
\label{fig:appendix_imgnet_power_curve}
  \caption{Error-cost plots for ImageNet-16H data with $N=3,6$. Fixed baselines \xis and \xfs are also included and depict strong but variable results. By contrast, \is and \fs maintain consistent results across budget and hyperparameter settings.}
\end{figure}

\paragraph{Results: Distribution Shift (Section 4.4 in paper)} %
\label{par:dist_shift}

\begin{figure}[h!]
\begin{tabular}{cc}
  \includegraphics[width=0.49\textwidth]{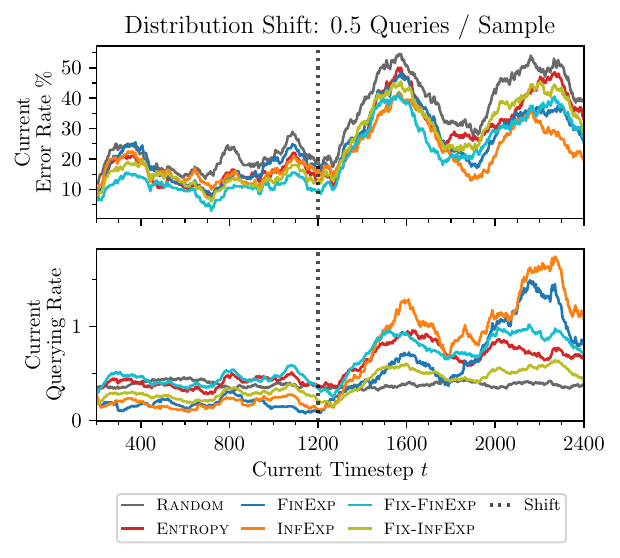} &   \includegraphics[width=0.49\textwidth]{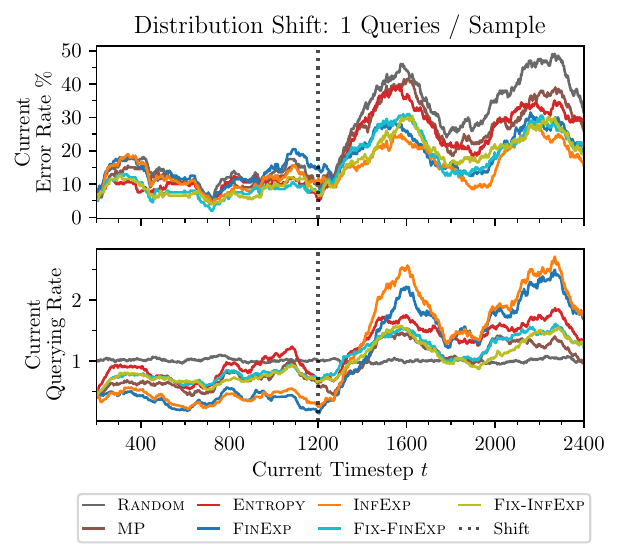} \\
    \includegraphics[width=0.49\textwidth]{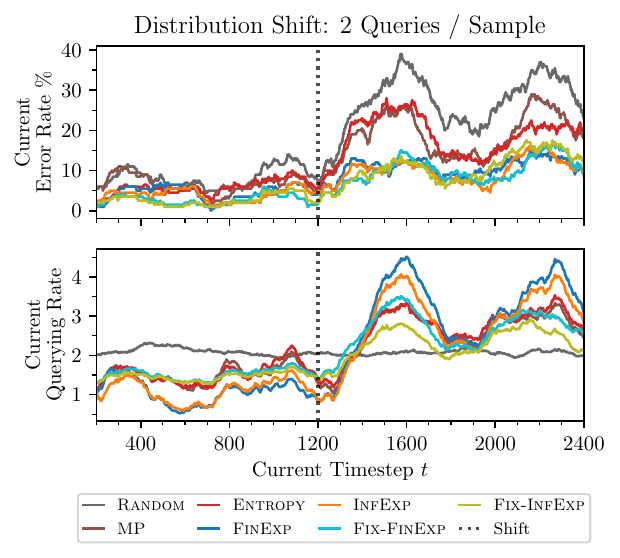} &   \includegraphics[width=0.49\textwidth]{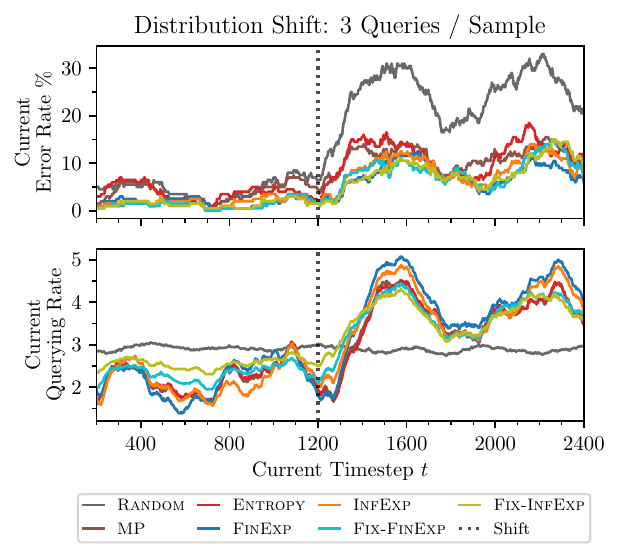} \\
\end{tabular}
\caption{Distribution shift analysis for all methods and baselines across different querying budgets on ImageNet-16H, with performance converging at high budgets when consensus is reached on most samples.}
\label{fig:dist_shift_budget}
\end{figure}

To ablate our distribution shift findings in \cref{fig:dist_shift_budget}, we report a selection of results for all methods in different budget settings. We see consistent results across these experiments. For the same querying budget, our proposed methods consistently offer improved performance under distribution shift as well as increased sensitivity to the shift as measured by adjusted querying budgets. That is, our methods tend to under-query baselines pre-shift, but over-query post-shift. This behavior demonstrates the promise of these methods to be robust under distribution shifts in real-world situations. At the highest budgets (i.e. 3 queries / sample), we naturally see the methods converge in budget and performance; such a high budget on ImageNet-16H with $N=6$ often yields absolute ground truth. Even so, in these settings we still consistently (slightly) outperform the baselines, as shown in \Cref{tab:dist_shift_appendix}. 

\begin{figure}[h!]

\begin{tabular}{cc}
  \includegraphics[width=0.48\textwidth]{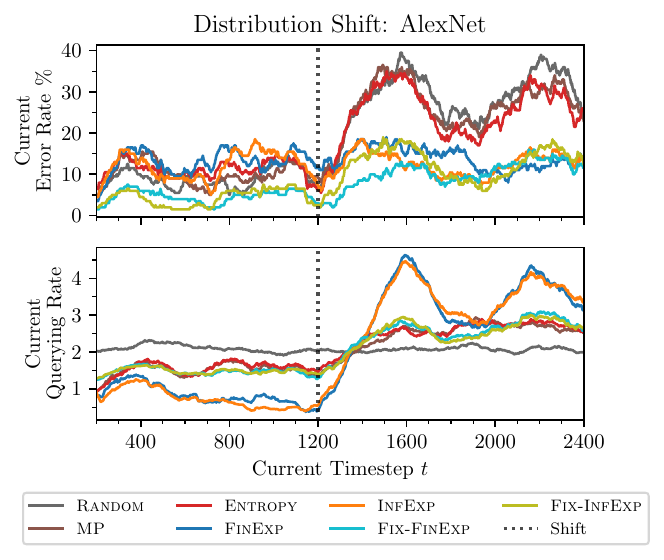} &   \includegraphics[width=0.48\textwidth]{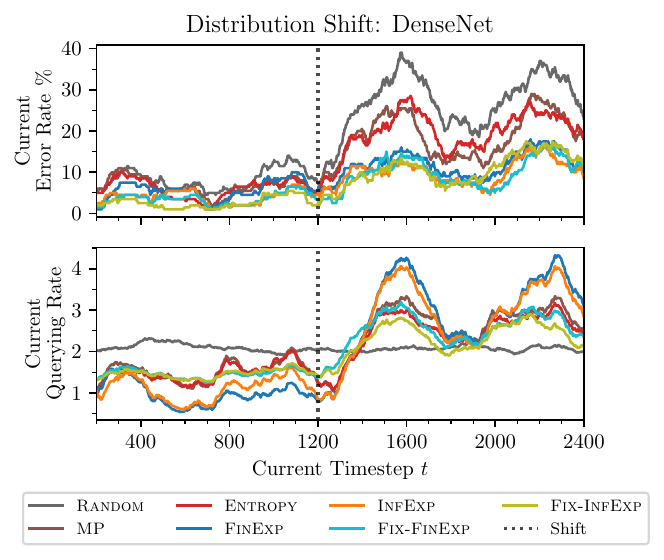} \\
    \includegraphics[width=0.48\textwidth]{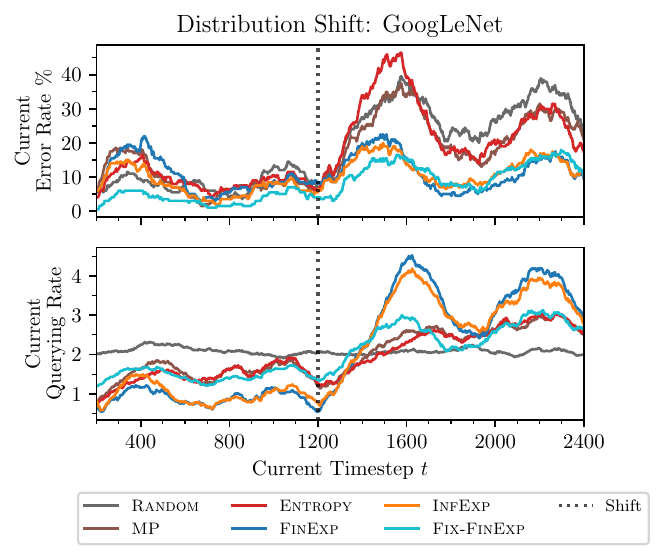} &   \includegraphics[width=0.48\textwidth]{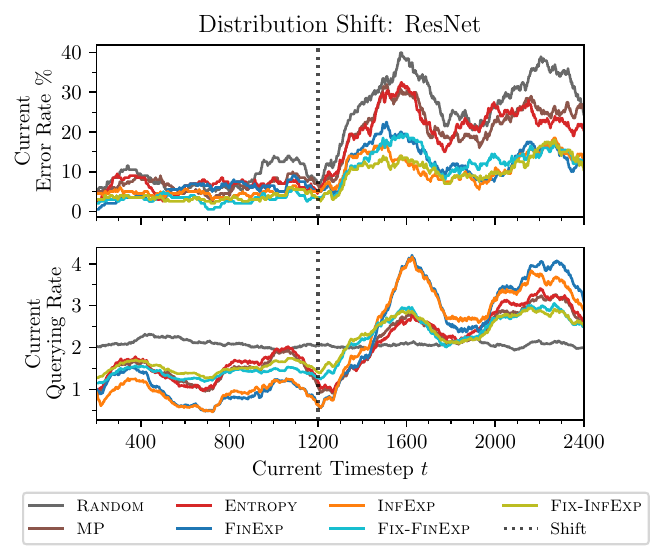} \\
    \multicolumn{2}{c}{\includegraphics[width=0.48\textwidth]{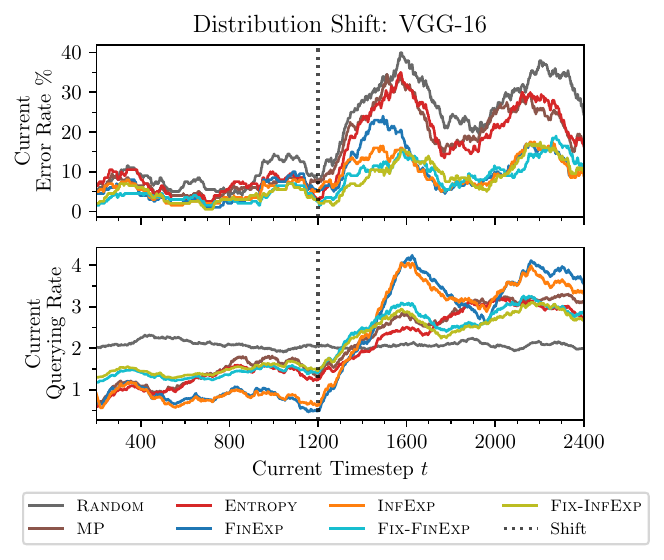}} \\
\end{tabular}
\caption{Distribution shift architecture ablation where runs with an average querying budget of 2 queries / sample are compared across five architectures on ImageNet-16H.}
\label{fig:dist_shift_arch}
\end{figure}

To validate our findings remain consistent across different model architectures, we select a budget of 2 queries / sample and evaluate the performance of methods across architectures, visualized below in \Cref{fig:dist_shift_arch}. Findings remain consistent with \Cref{fig:dist_shift_budget} and the results in the main paper. However, we note that fixed prior methods \xfs and \xis do not demonstrate the same budget sensitivity to distribution shifts and possess similar querying budget behavior to the baselines.

Across all runs, we note in \Cref{tab:dist_shift_appendix} a significant and consistent improvement in error rate for all budgets with our proposed methods. These findings hold even in cases where the pre-shift error rates of all methods are comparable, such as with a querying budget of 0.5 queries / sample.

\begin{table*}[h]
\centering
\fontsize{6pt}{6pt}\selectfont
\caption{Error rate (\%) of all methods before and after distribution shift on ImageNet-16H, averaged over model architectures and grouped by approximate querying budget per sample (0.5, 1, 2, 3). }
\begin{tabular}{l|rr|rr|rr|rr}
\toprule
Queries / Sample&  \multicolumn{2}{c}{0.5} &  \multicolumn{2}{c}{1} & \multicolumn{2}{c}{2} & \multicolumn{2}{c}{3}\\ 
  \toprule
 Method & 
 \multicolumn{1}{c}{Pre-shift} & \multicolumn{1}{c|}{Post-shift} & \multicolumn{1}{c}{Pre-shift} & \multicolumn{1}{c|}{Post-shift} & \multicolumn{1}{c}{Pre-shift} & \multicolumn{1}{c|}{Post-shift} & \multicolumn{1}{c}{Pre-shift} & \multicolumn{1}{c}{Post-shift} \\
 \toprule
 \textsc{RANDOM}   & 20.23±1.26 & 47.55±2.34 & 13.21±0.29 & 40.03±0.48 & 7.9±0.11 & 29.28±0.24 & 4.95±0.05 &  25.5±0.03  \\ 
 \textsc{ENTROPY}   & 15.02±0.52 & 38.85±0.69 & 9.38±0.41 & 28.58±0.56 & 4.02±0.12 & 13.66±0.27 & 2.52±0.08 &  9.10±0.09  \\ 
 \textsc{MP}   & \textbf{11.26±0.35} & 31.65±0.9 & 9.06±0.33 & 28.0±0.64 & 3.85±0.12 & 13.13±0.26 & 2.44±0.06 &  9.03±0.07  \\ 
 \midrule
 \fx   & 11.56±0.88 & 32.27±0.97 & 5.63±0.55 & 19.66±1.16 & 1.63±0.06 & \textbf{8.25±0.16} & 1.25±0.07 &  7.83±0.14  \\ 
 \fs   & 13.07±1.15 & 21.14±1.63 & 5.84±0.31 & 11.98±0.19 & 2.18±0.11 & \textbf{8.26±0.13} & 1.31±0.05 &  \textbf{7.24±0.13}  \\ 
 \is   & 11.55±0.23 & \textbf{19.65±0.25} & \textbf{5.47±0.19} & \textbf{11.52±0.17} & \textbf{1.54±0.09} & \textbf{8.26±0.20} & \textbf{1.19±0.07} &  7.68±0.16  \\ 
  \bottomrule                          
\end{tabular}
\label{tab:dist_shift_appendix}
\end{table*}

\paragraph{Results: Time-averaged Distribution Shift}

Across several of our distribution shift ablation experiments we noted several fluctuations in the querying and error rate, particularly post-shift. We investigated this phenomena and discovered that this was a product of the specific sequence orderings we were averaging over and the moving average smoother accentuating these fluctuations. 

\begin{figure}
    \centering
    \includegraphics{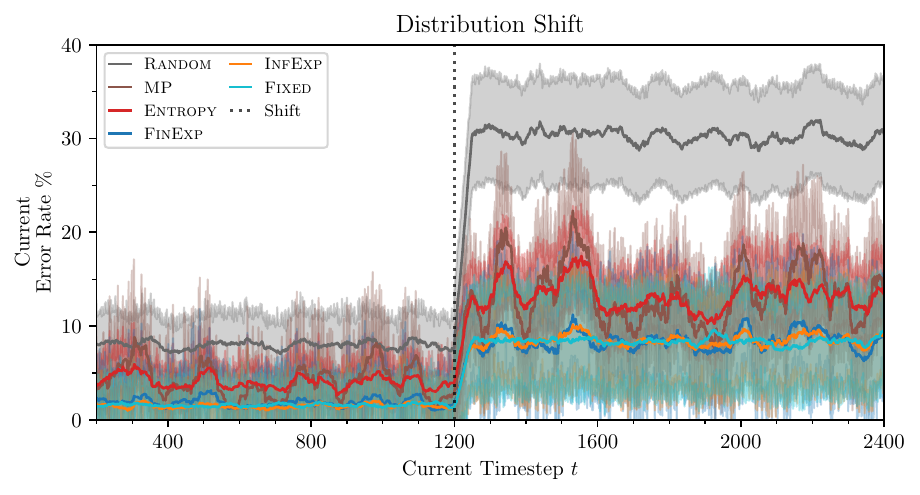}
    \caption{Time-averaged error-rate over 100 sequences of proposed methods and baselines for distribution shift plots of ImageNet-16H. All methods are set to consume a querying budget of roughly 2 queries / sample. Hue represents roughly a 95\% confidence interval as measured by the standard error. }
    \label{fig:agg_dist_shift}
\end{figure}

Therefore, we conducted an additional analysis by running distribution shift experiments over 100 different random sequence orderings and averaging the error rate performance. This offers us two insights. First, it assists in validating the performance of our proposed methods over time is superior two the baselines. In addition, we can investigate a time-averaged plot to attempt to observe any distribution shift convergence behavior. Below in \cref{fig:agg_dist_shift}, we note that our proposed methods consistently outperform the baselines and possess smaller fluctuations in error rate. We attribute the larger error hue not to the performance of our methods but rather to the variance in difficulty in classifying specific samples. For this reason, and likely due to the relatively small number of samples we evaluate on, we are not able to identify any convergence behavior, where the model adjusts to the distribution shift and reaches a ``steady-state'' error rate. This aligns with our intuition on two-phase evaluation results in \cref{tab:appendix_two_phase} which demonstrate that our learned prior parameters can only recalibrate the existing model $f$ enough to yield an error rate reduction of a few percentage points.

\paragraph{Results: Model Calibration Analysis (Section 4.5 in paper)} %
\label{par:model_calibration}

To further analyze the ability for our framework to learn model performance and adjust predictions accordingly, we extend the visual findings of \cref{fig:tau_analysis} in the main paper by aggregating many runs and grouping them across budgets. In \cref{tab:appendix_two_phase} below, we present the average Pearson correlations between the learned parameters $\tau$ and the per-class model performance relative to consensus. Consistently, we witness strong positive correlations in these values across all budgets and model performance levels. Increased querying budget generally tends to also increase this correlation, though this finding does not always persist across settings.

\begin{table}[h!]
\centering
\fontsize{8pt}{8pt}\selectfont
\caption{Pearson correlation between learned MAP parameters $\tau$ and per class performance of $f$.}
\begin{tabular}{cc|ccccc}
\toprule
\multicolumn{2}{c}{CIFAR-10H}& \multicolumn{4}{c}{Avg. Queries per Sample} \\ 
\midrule
$f$-Acc. &Method & 0.5 &  1& 2 & 3 \\ 
\midrule
 &\textsc{Fix-InfExp}   & 0.550±0.024 & - & 0.614±0.004 & 0.712±0.039 \\ 
\multirow{2}*{50\%}&\textsc{Fix-FinExp}   & 0.625±0.060 & - & 0.477±0.002 & 0.471±0.008 \\ 
 &\textsc{InfExp}   & 0.576±0.007 & 0.638±0.012 & 0.649±0.009 & 0.562±0.034 \\ 
&\textsc{FinExp}   & 0.622±0.014 & 0.623±0.020 & 0.560±0.012 & 0.590±0.038 \\ 
\midrule
& \textsc{Fix-InfExp}   & - & 0.600±0.016 & 0.662±0.002 & 0.670±0.020 \\ 
\multirow{2}*{70\%} & \textsc{Fix-FinExp}   & - & 0.573±0.013 & 0.577±0.002 & 0.506±0.005 \\ 
& \textsc{InfExp}   & 0.576±0.01 & 0.640±0.012 & 0.712±0.006 & 0.705±0.011 \\ 
& \textsc{FinExp}   & 0.600±0.008 & 0.601±0.015 & 0.662±0.013 & 0.640±0.019 \\ 
\midrule
& \textsc{Fix-InfExp}   & - & 0.669±0.021 & 0.702±0.002 & - \\ 
\multirow{2}*{90\%} & \textsc{Fix-FinExp}   & - & 0.687±0.014 & 0.590±0.002 & - \\ 
& \textsc{InfExp}   & 0.521±0.011 & 0.587±0.006 & 0.712±0.013 & 0.767±0.024 \\ 
& \textsc{FinExp}   & 0.589±0.005 & 0.681±0.018 & 0.659±0.041 & 0.672±0.022 \\    
  \bottomrule                          
\end{tabular}
\label{tab:appendix_two_phase}
\end{table}